\documentclass[sn-mathphys-num,iico]{sn-jnl}

\usepackage{graphicx}%
\usepackage{multirow}%
\usepackage{booktabs}%
\usepackage{amsmath,amssymb,amsfonts}%
\usepackage{amsthm}%
\usepackage{mathrsfs}%
\usepackage[title]{appendix}%
\usepackage{xcolor}%
\usepackage{textcomp}%
\usepackage{manyfoot}%
\usepackage{booktabs}%
\usepackage{algorithm}%
\usepackage{algorithmicx}%
\usepackage{algpseudocode}%
\usepackage{listings}%
\definecolor{LightCyan}{rgb}{0.88,1,1}
\definecolor{mygreen}{RGB}{28,172,0} 
\definecolor{mylilas}{RGB}{170,55,241}
\usepackage{subcaption}
\usepackage{bm}
\usepackage{tikz}
\usetikzlibrary {arrows.meta,bending,positioning}
\usetikzlibrary{patterns}
\usepackage{matlab-prettifier}
\usetikzlibrary{shapes.multipart}
\usetikzlibrary{tikzmark}
\usepackage[utf8]{inputenc}
\usepackage{flushend}
\usepackage{tikz,pgfplots}
\usepackage{blkarray}
\usetikzlibrary{calc}
\usetikzlibrary{decorations.pathreplacing,angles,quotes}
\usetikzlibrary {decorations.fractals,spy}
\usepackage{tikz-dimline}
\usepackage{siunitx}

\begin{document}

\title{ Flap like a bat: Soft pneumatic wings for biomimetic ornithopters}

\author[1]{\fnm{Besetty Satyanarayana} \sur{Pardhu}}
\author[1]{\fnm{Sirivelu} \sur{Akhilesh}}
\author[1]{\fnm{Bammidi} \sur{Abhilash}}
\author*[1]{\fnm{Prabhat} \sur{Kumar}}\email{pkumar@mae.iith.ac.in}

\affil*[1]{\orgdiv{Mechanical and Aerospace Engineering}, \orgname{Indian Institute of Technology Hyderabad}, \orgaddress{\street{Sangareddy}, \city{Hyderabad}, \postcode{502285}, \state{Telangana State}, \country{India}}}


%
\abstract{
	This paper describes the development of a bio-inspired ornithopter wing mechanism that uses soft pneumatic actuators, such as PneuNet structures, in place of revolute joints, inspired by the complicated kinematics of bat flight. The proposed design employs pneumatically driven air pockets within a flexible wing skeleton to achieve smooth, lifelike flapping motion. The actuation system is controlled by 3/2-way solenoid valves and an air pump, which manipulate airflow and actuator deformation. We demonstrate, using ANSYS FEA software, the bending behavior of the PneuNet actuators under varying pressures to inform the wing's structural design. The implementation of soft robotics in the flapping-wing system offers potential advantages over traditional mechanisms, such as reduced mechanical complexity, lower acoustic signatures, enhanced adaptability, and wing morphing. This work demonstrates the feasibility of replacing rigid revolute joints with compliant pneumatic actuators, enabling more agile, silent, and biomimetic aerial robots. 
}
\maketitle
\section{\textbf{Introduction}}
\begin{figure}[h]
	\centering
	\begin{subfigure}{0.49\textwidth}
		\centering
		\includegraphics[scale=0.2]{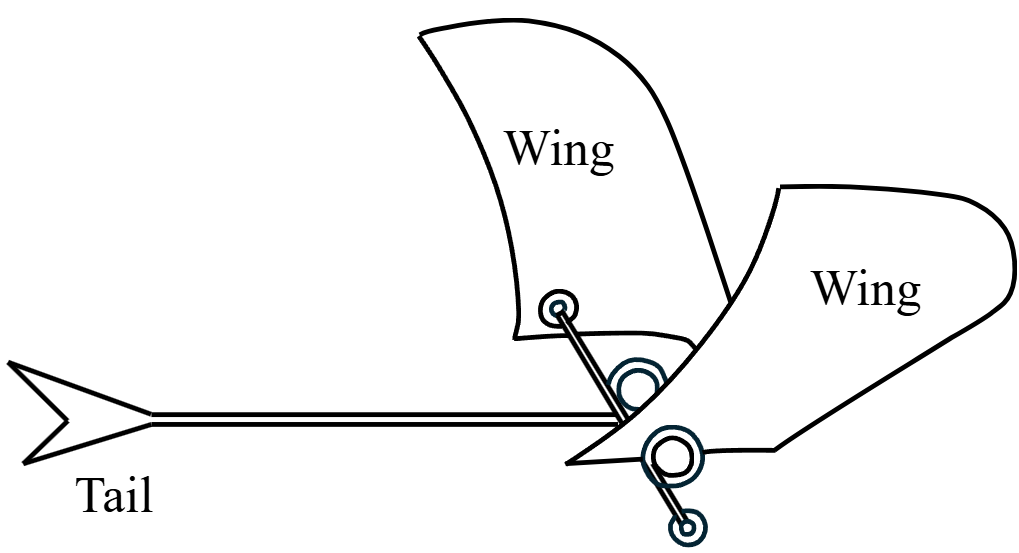} 
		\caption{A schematic of human powered ornithopter design} \label{fig:daVinci}
	\end{subfigure}
	\begin{subfigure}{0.49\textwidth}
		\centering
		\includegraphics[scale=0.90]{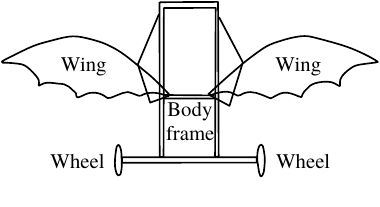} 
		\caption{A schematic of 19th century ornithopter prototype}
		\label{fig:edwardFrost}
	\end{subfigure}
	\caption{Schematic diagrams inspired by early ornithopter concepts}
\end{figure}

\begin{figure}[h]
	\centering
	\begin{subfigure}{0.49\textwidth}
		\centering
		\includegraphics[scale=0.60]{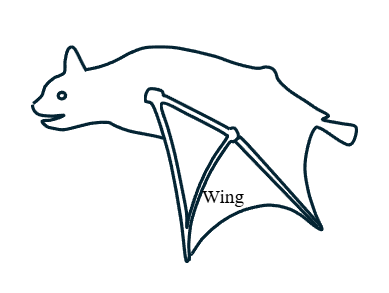} 
		\caption{Mid-downstroke} 
	\end{subfigure}
	\begin{subfigure}{0.49\textwidth}
		\centering
		\includegraphics[scale=0.55]{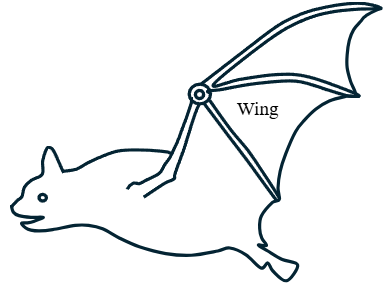} 
		\caption{Mid-upstroke}
	\end{subfigure}
	\caption{Schematic of active wing retraction during flight} \label{fig:batRetraction}
\end{figure}

The evolution of ornithopters, flying machines that mimic the flapping-wing motion of birds and bats, has spanned centuries due to the curiosity of how birds and bats fly. Early conceptualizations on the biomechanical understandings of bird wings can be traced back to Leonardo da Vinci's 15th-century sketch (Fig.~\ref{fig:daVinci}), where he envisioned flying wings powered by humans ~\cite{daVinciCodex}. Although never realized in his time, these ideas laid the philosophical foundation for biomimetic flight.
The 19th and early 20th centuries saw many attempts at building mechanically flapping aircraft using wood, fabric, and crank mechanisms (Fig.~\ref{fig:edwardFrost}). These models were large, inefficient, and incapable of long flight. Small-scale ornithopters could be made with the introduction of lightweight materials and electric motors in the mid-20th century. Researchers began using rotary motors with cam-driven or kinematic linkage-based flapping mechanisms, enabling basic yet repeatable wing motion (Fig.~\ref{fig:batRetraction})~\cite{batRetraction}. Modern designs were further advanced with servo motors and microcontrollers, to make programmable wing kinematics much easier and improve flight stability~\cite{wood2008insect}. Despite these technological advances, traditional ornithopters relied solely on rigid mechanical components, such as revolute joints and fixed kinematic linkages. These systems introduce several issues: they produce significant mechanical noise due to oscillations and an unstable center of mass, are mechanically complex, and lack the compliance and adaptability seen in biological flyers~\cite{majidi2013perspective}. Among biological flyers, bats stand out due to their highly articulated wings supported by flexible membranes and independently actuated joints. This allows for better morphing of wing shape, surface area, and curvature during flight; these characteristics contribute to their exceptional maneuverability and energy efficiency~\cite{ramezani2017bat,miyashita2021aerobat}.

\begin{figure}
	\centering
	\includegraphics[scale = 0.65]{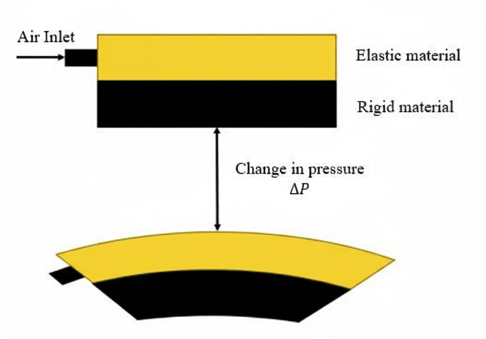}
	\caption{A schematic diagram for a PneuNet actuator}
	\label{fig:PneuNetdesign}
\end{figure}

To achieve these features, this study proposes a soft robotic approach: the use of PneuNet actuators~\cite{shepherd2011multigait}. These flexible pneumatic networks bend if we change the internal air pressure (Fig.~\ref{fig:PneuNetdesign}), to replace traditional revolute mechanisms~\cite{kumar2022towards}. These actuators are lightweight, compliant, and silent, making them ideal candidates for any aerial vehicle with revolute mechanisms~\cite{kumar2022towards}. The flapping mechanism is actuated by a pneumatic control system comprising 3/2-way solenoid valves and an air pump that draws air from the external atmosphere. We used ANSYS to perform the FEA simulations to understand the deformation behavior of the actuators under different pressure conditions. The design and initial modeling are performed using open-source PneuNet tools from MIT's Soft Robotics Toolkit~\cite{softroboticsMIT}. 

The primary goals of this study are outlined below:
\begin{itemize}
	\item To design a bat-inspired flapping wing integrated with soft pneumatic actuators
	\item To simulate actuator performance using FEA, and
	\item To evaluate the feasibility and advantages of soft actuation in flapping-wing flight systems.
\end{itemize}
The remainder of the paper is structured as follows. Sec.~\ref{Sec:LitR} provides a state-of-the-art of the topic. The design and methodology are provided in Sec.~\ref{Sec:DesignMethodlogy}. Sec.~\ref{Sec:SimulTest} provides the ANSYS simulation steps and results. Lastly, closing remarks are noted in Sec.~\ref{Sec:Closing}.

\section{Literature Review}\label{Sec:LitR}
Recent advances in soft robotics and biomimetic design have enabled innovative approaches to the design of flapping-wing mechanisms. Traditional ornithopters mostly use rigid kinematic linkages and servo motors, which, while effective for basic flapping motion, lack the flexibility and lifelike articulation seen in bats. 

\begin{figure}[h!]
	\centering
	\includegraphics[width=0.6\linewidth]{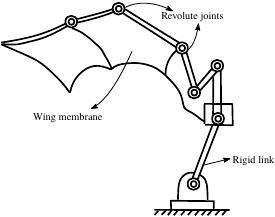}
	\caption{Schematic diagram of a bat-bot}\label{fig:bat-bot}
\end{figure}

Key research by Ilievski et al.~\cite{ilievski2011soft} and Polygerinos et al.~\cite{polygerinos2015modeling} introduced and refined PneuNet actuators, showing that PneuNets can generate controlled, compliant motion using pneumatic inflation. These actuators have been widely used in soft grippers and walking robots, but have seen limited use in aerial systems. Soft robotics literature, such as Laschi and Cianchetti~\cite{laschi2014soft} and Chiang et al.~\cite{chiang2021flapping}, highlights the potential of soft actuated joints in adaptable, low-noise locomotion—principles that are perfect for winged mechanisms like ornithopter flight. In the study of bat-inspired aerial robots, notable examples include Bat Bot~\cite{ramezani2017biomimetic} (Fig.~\ref{fig:bat-bot}) and AeroBat~\cite{miyashita2021aerobat}. These designs attempted to replicate bat wing motion using articulated mechanical systems, but still rely heavily on servo motors and rigid kinematic linkages (Fig.~\ref{fig:bat-bot}).

Bat wings have long arms and fingers attached to a flexible skin membrane (patagium). This helps achieve multiple degrees of freedom, allowing independent joint control and active wing morphing during flight. Bats extend their wings fully to generate lift during the downstroke, and then retract them close to their body during the upstroke to reduce drag (Fig.~\ref{fig:batRetraction}). This improves agility, fine maneuvering, and aerodynamic efficiency in restricted environments with many obstacles. Despite these advantages, limited studies have been conducted on the morphing wing mechanism using pneumatic actuators. This study aims to assess the feasibility of using PneuNets in winged mechanisms to enhance realism, stealth, and maneuverability in aerial robotics.

\section{Design and Methodology}\label{Sec:DesignMethodlogy}
This section presents the design and methodology of the proposed pneumatically actuated air pockets integrated within a flexible wing skeleton for bat-inspired flight.
\subsection{Skeleton Structure}
\begin{figure}
	\centering
	\includegraphics[width=0.8\linewidth]{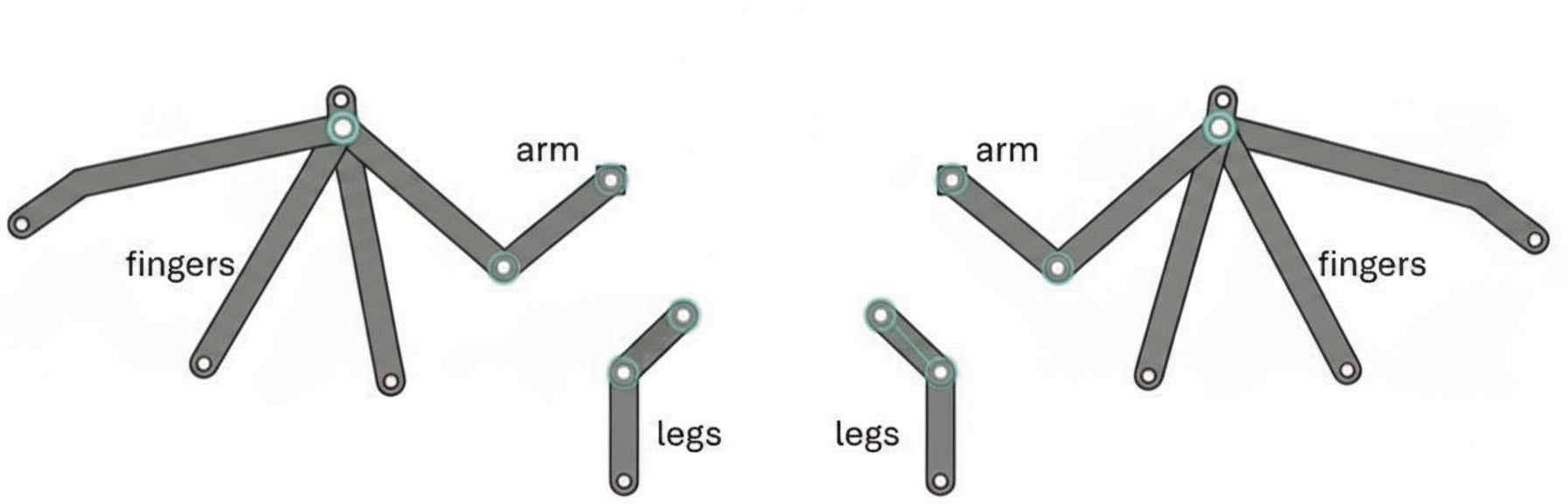}
	\caption{Skeleton structure of bat-inspired ornithopter}
	\label{fig:Skelt}
\end{figure}

\begin{figure}
	\centering
	\includegraphics[width=0.6\textwidth]{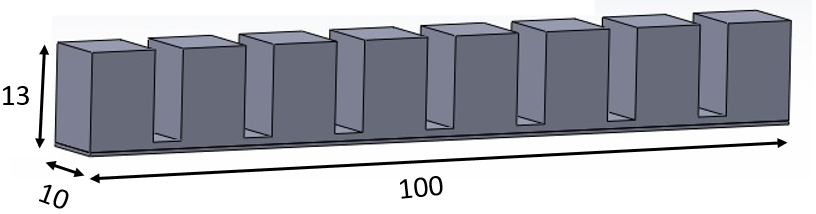} 
	\caption{A CAD model of PneuNet actuator (dimensions are in mm)}
	\label{fig:ourPrototype}
\end{figure}
The structural design of the ornithopter is based on the anatomy of a bat's wing. The upper parts of Fig.~\ref{fig:Skelt} are the arms and fingers of the wing. The lower links are the bat's legs, and a flexible material is used to connect the ends of the fingers and legs to the body, similar to the patagium in bats. This helps in increasing the area of the wing used to generate lift. 

All these joints in the structure are replaced with PneuNet actuators. These actuators independently control every angle between the joints. During the upstroke, all angles between the linkages will be minimized to reduce drag force, and during the downstroke, angles will be returned to normal to maximize the wing's area and increase lift by pushing more air. 

The manufacturing process is also designed to be practical. The skin can be produced using either the rolling or calendering method. The actuators themselves are made using stereolithography printing with silicone materials. The bones can be 3D-printed or manufactured using subtractive techniques, providing flexibility in the structure's design.  

\subsection{PneuNet Actuator}
The PneuNet actuator is a soft robotic component that replicates biological joint movements by combining pneumatic actuation with hyper-elastic materials. The CAD model of the PneuNet actuator, as shown in Fig.~\ref{fig:ourPrototype}, consists of a series of cuboidal chambers aligned along a compliant base. In this design, the cuboidal chambers are hollow, allowing pressurized air to be injected. These chambers are fabricated using Silicone 40A, a hyper-elastic elastomer known for its high elasticity, low stiffness, and excellent flexibility. When air is introduced into the chambers, the silicone walls expand, enabling the structure to deform and perform mechanical work through pneumatic inflation.

The actuator is equipped with a base layer fabricated from nylon fabric to direct this expansion into a predictable bending motion. This material, which is bio-compatible and optimized for medical and soft robotics applications, exhibits a moderate Young's modulus (1-2 GPa), significantly higher than that of Silicone 40A, but still flexible enough to allow bending while restricting in-plane stretch. Unlike rigid plastics, nylon fabric maintains compliance while acting as an inextensible constraint layer, preventing the base from elongating axially during pressurization.

When the chambers are inflated, the silicone 40A walls tend to expand omnidirectionally due to their hyper-elastic nature. However, because the nylon fabric base resists axial elongation, the deformation is forced to propagate as a curved bending motion. This mechanism converts uniform chamber expansion into a directional bending of the entire actuator. The degree of bending can be finely controlled by regulating the input pressure, providing an adjustable actuation system well-suited for soft robotics.

In this research, the PneuNet actuator serves as a replacement for conventional revolute joints in an ornithopter-inspired flapping mechanism. The actuators are connected to the artificial "bones" of the ornithopter's skeletal structure, functioning as compliant, soft joint actuators. Unlike traditional rigid mechanisms, this approach allows the wing to flap with bio-inspired smoothness and compliance, closely mimicking the joint motion of natural flyers such as bats. Also, both materials are compatible with accessible manufacturing processes, including 3D printing and molding, making this approach practical for research and development in soft robotics and bio-mimetic systems.

\subsection{Pneumatic System}
\begin{figure}[h!]
	\centering
	\includegraphics[width=0.8\linewidth]{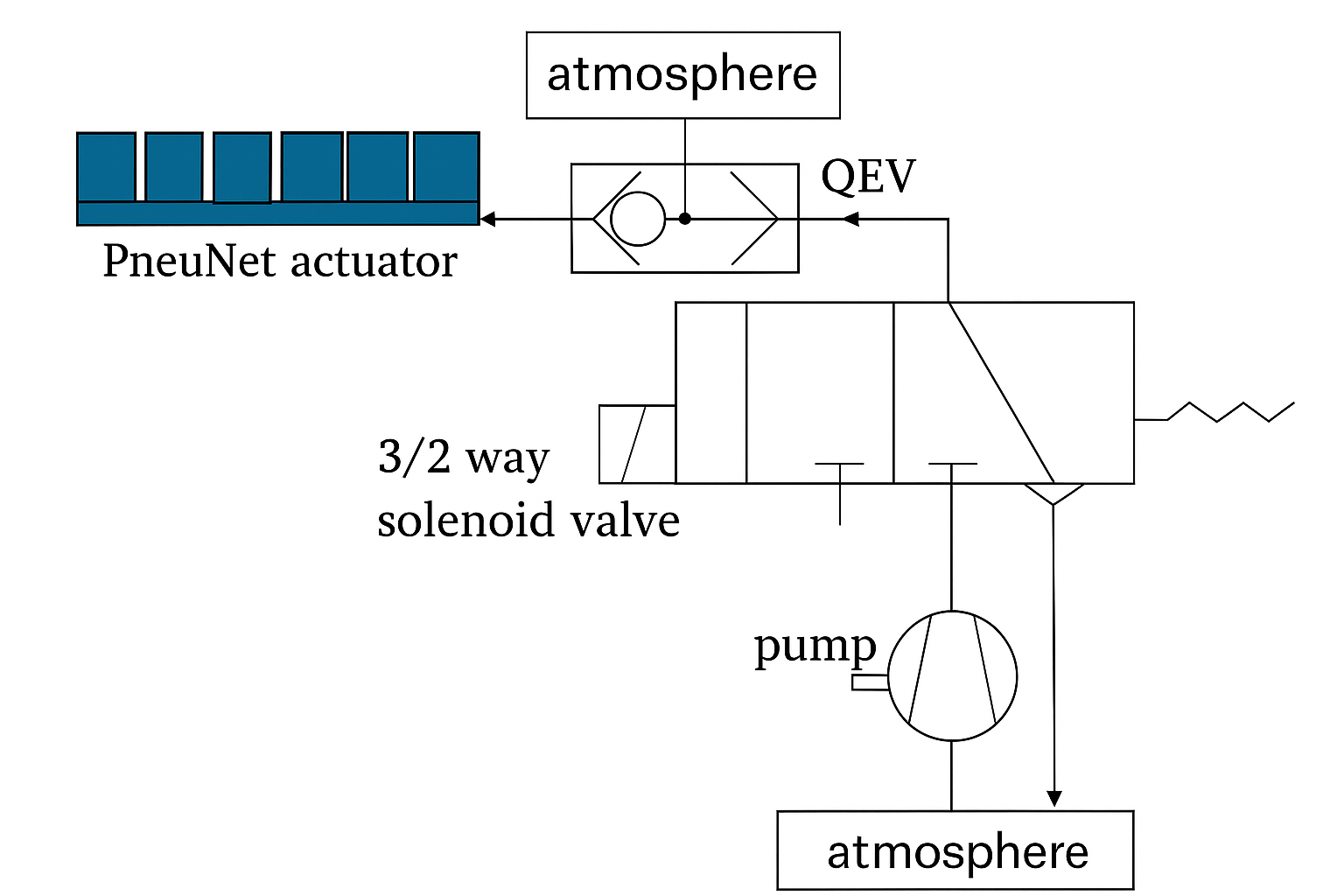}
	\caption{Actuation flow chart}
	\label{fig:Flowchat}
\end{figure}

The PneuNet actuator serves as a flexible joint in place of traditional rigid components. It is built from soft elastomeric materials that expand and bend when air is pumped into them. These actuators are designed to bend in a controlled way, much like natural joints found in bats. Compressed air is supplied to the actuator via a solenoid valve, while a quick exhaust valve (QEV) is used for rapid deflation.

When the ESP32 microcontroller switches on the solenoid valve, compressed air flows into the actuator (Fig.~\ref{fig:Flowchat}). The internal pressure causes the actuator walls to expand, and because the base material resists stretching in one direction, the expansion results in a smooth bending motion. The actuator continues to bend until the air pressure is stopped. Once the solenoid is turned off, the input pressure drops suddenly. The QEV then quickly deflates the PneuNet actuator. This quick release allows the actuator to return to its original shape quickly. By controlling solenoid valves, the bending and relaxing of each joint can be controlled. This can help control the wing joints independently, mimicking the way real bats fly. The materials used are compatible with standard 3D printing and molding techniques, making this design approach efficient and accessible for bio-inspired robotic systems. The entire actuation system is based on pneumatic control rather than rigid mechanics. At the center of this system is a 3/2-way solenoid valve with an air input, an air output, and an exhaust. The input is connected to a diaphragm air pump that draws in ambient air. The Output sends compressed air to the PneuNet actuator and the Quick Exhaust Valve. The exhaust is connected to the atmosphere. This setup allows continuous air pumping, with flow controlled by switching the solenoid. When air is directed into the actuator, the QEV stays closed, and the actuator inflates. When the solenoid turns off, the inlet pressure drops, and the QEV opens immediately, venting the actuator. This rapid inflation-deflation cycle allows the wings to flap at 6–7 Hz. To achieve such fast, responsive motion, the solenoid valves and QEVs must be placed as close as possible to the actuators to minimize dead volume in the tubing.

The airflow is drawn from near the body to avoid affecting the wings' aerodynamics. The control sequence is simple: all solenoids are activated to fold the wings during the upstroke. The solenoids are turned off during the downstroke, letting the wings open. A slight delay is added to the solenoid between the wing and the body to ensure that flapping begins only after the wing is fully extended. For reliable high-speed switching, the system uses SMC V100 solenoid valves operating at 20–40 Hz, along with an ESP32 and MOSFETs for signal control. The valves are operated within 8–10 milliseconds, allowing precise timing for every wingbeat. 

\begin{figure}
	\centering
	\includegraphics[width=0.8\linewidth]{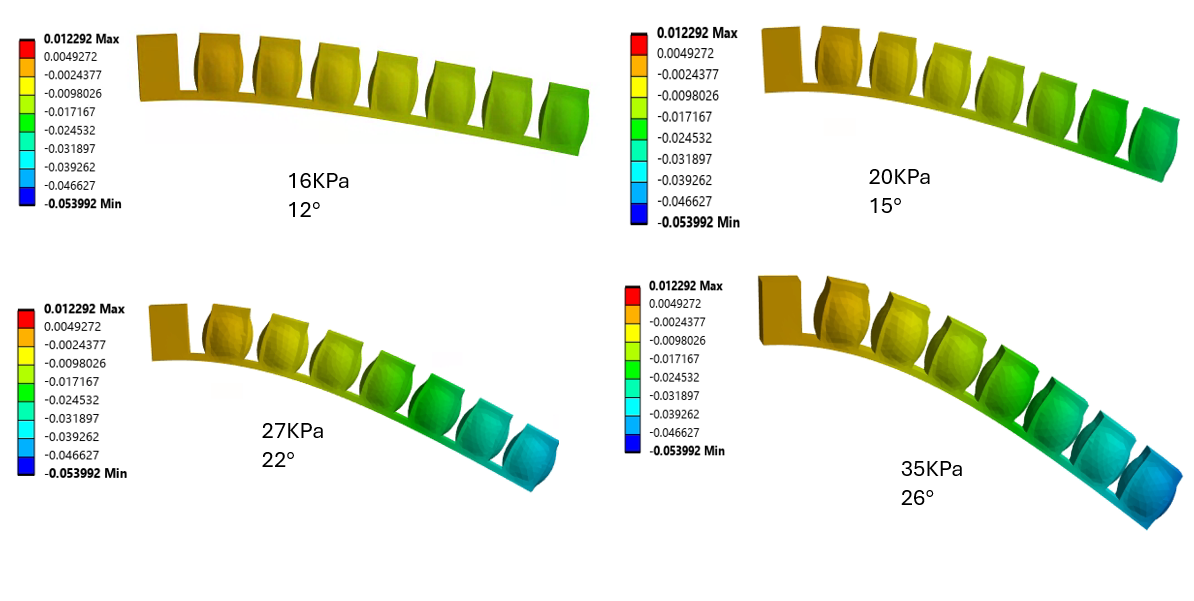}
	\caption{Simulation results}
	\label{fig:deformationProfiles}
\end{figure}

\begin{figure}
	\centering
	\begin{tikzpicture}
		\pgfplotsset{compat = 1.3}
		\begin{axis}[ blue,
			width = 0.75\textwidth,
			xlabel=Pressure (kPa),
			axis y line* = left,
			ylabel= Bending angle (in degrees),
			ymajorgrids=true,
			xmajorgrids=true,
			grid style=dashed,
			legend style={at={(0.50,0.75)},anchor=east} ]
			\pgfplotstableread{bend.txt}\mydata;
			\addplot[black, mark = square,mark size=1pt,style={very thick}]
			table {\mydata};
		\end{axis}
	\end{tikzpicture} 
	\caption{Bending angle}
	\label{fig:Bendingangle}
\end{figure}
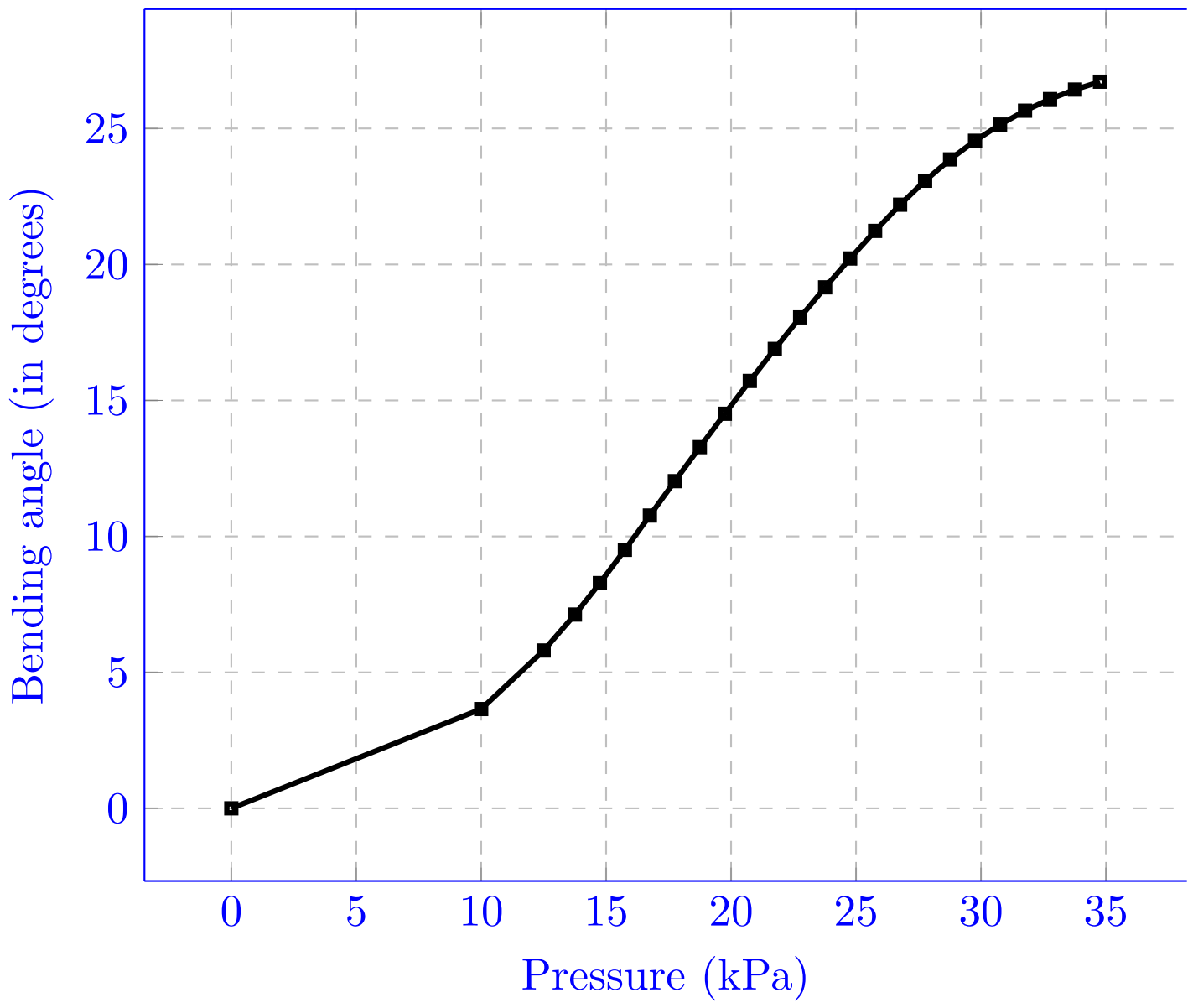 
\section{Simulation and Testing}\label{Sec:SimulTest}

In this work, we used silicone elastomer (shore hardness 40A) as the stretchable material for the actuator body, as it can be 3D-printed using soft lithography molds. The hyper-elastic behavior of silicone 40A was modeled using the Mooney Rivlin 2-parameter model using material constants, $C10 = 0.085$ MPa, $C01 = 0.015$MPa, and D= $\SI {1e-4}{\per \mega \pascal}$. These parameters were chosen based on known physical behavior. C10 governs low-strain stiffness and controls how easily the actuator begins to deform. C01 contributes to the material's response at high strain and captures nonlinearity. D governs compressibility, with a small value ensuring silicone's nearly incompressible nature. High values of C10+C01 indicate high material stiffness and minimal deformation, whereas too low values can lead to large deformations. A low D value is essential to avoid unrealistic volume changes during inflation. 

To ensure directional deformation, the actuator's bottom layer was modeled as an inextensible base constructed from nylon fabric. Nylon exhibits high stiffness (Young's modulus of 1–2 GPa) and was modeled as a linear elastic material. This inextensible layer restricts expansion on one side, ensuring that internal pressure induces asymmetric deformation and produces controlled bending motion. 

The CAD model (Fig.~\ref{fig:ourPrototype}) was constructed using the two material regions described above. During simulation, one end of the actuator was fixed, replicating the attachment point to a rigid link in the bat-inspired skeleton structure. Pressurization of the internal chambers helps in directional bending deformation, enabling actuation. This configuration would allow dynamic flapping motion by alternating the pressurization and depressurization cycles, mimicking wing strokes. This simulation replicates the working principle of a PneuNet actuator, in which internal pressure drives motion through the differential material responses of the stretchable and inextensible layers. As the fabric covering the skeletal structure also has some elasticity, which acts like a spring, the actual pressure vs bending plot will differ slightly. The results in Fig.~\ref{fig:deformationProfiles} show how the PneuNet actuator bends by applying increasing pressure. The bending angle of the PneuNet is depicted in Fig.~\ref{fig:Bendingangle}

\section{Closing remarks and future work} \label{Sec:Closing}
This research presents an innovative approach to ornithopter wing design by employing soft-robotic PneuNet actuators to mimic the flapping mechanics of bat wings, unlike traditional rigid ornithopters, which use servo motors and kinematic linkages. This system primarily leverages the compliant, flexible nature of soft robotics to more accurately replicate the morphing wing dynamics of bats, which retract their wings during the upstroke to minimize drag and enhance energy efficiency. 

The use of soft pneumatic actuators not only enables more biomimetic movement but also addresses key limitations of rigid designs, such as noise and mechanical complexity. Soft robotics introduces elements of safe interaction with the environment and lightweight construction, which are critical for aerial robotics and for operation in restricted environments.

Therefore, the proposed bat-inspired soft-wing mechanism holds significant promise for aerial surveillance, search-and-rescue operations, environmental monitoring, and covert reconnaissance, where silent flight and high maneuverability are essential. In the case of any feasible response, this mechanism can be used quite often in the next-generation ornithopters that are quieter, safer, and more adaptable to real-world missions than their rigid-wing counterparts~\cite{daVinci}. A future avenue includes optimizing the design using a systematic topology optimization approach, fabricating, and testing the bat ornithopter. In addition, customized solenoids can be made using the sound dampers, thereby reducing the acoustic signature.



\end{document}